# Phase-fraction guided denoising diffusion model for augmenting multiphase steel microstructure segmentation via micrograph image-mask pair synthesis

Hoang Hai Nam Nguyen[1,2], Minh Tien Tran[1], Hoheok Kim[1], Ho Won Lee[1,2]*

[1] *Materials Data & Analysis Research Division, Korea Institute of Materials Science, Changwon, Republic of Korea*
[2] *Advanced Materials Engineering, University of Science and Technology, Daejeon, Republic of Korea*

**Abstract**

The effectiveness of machine learning in metallographic microstructure segmentation is often constrained by the lack of human-annotated phase masks, particularly for rare or compositionally complex morphologies within the metal alloy. We introduce *PF-DiffSeg*, a phase-fraction controlled, one-stage denoising diffusion framework that jointly synthesizes microstructure images and their corresponding segmentation masks in a single generative trajectory to further improve segmentation accuracy. By conditioning on global phase-fraction vectors, augmented to represent real data distribution and emphasize minority classes, our model generates compositionally valid and structurally coherent microstructure image and mask samples that improve both data diversity and training efficiency. Evaluated on the MetalDAM benchmark for additively manufactured multiphase steel, our synthetic augmentation method yields notable improvements in segmentation accuracy compared to standard augmentation strategies especially in minority classes and further outperforms a two-stage mask-guided diffusion and generative adversarial network (GAN) baselines, while also reducing inference time compared to conventional approach. The method integrates generation and conditioning into a unified framework, offering a scalable solution for data augmentation in metallographic applications.

*Ho Won Lee
*Contact :* +82-55-280-3843
*Email address :* h.lee@kims.re.kr

## 1. Introduction

Automated segmentation of metallic microstructures is pivotal in materials informatics, facilitating the quantification of morphology essential for structure–property modeling [1]. However, challenges such as the scarcity of pixel-accurate training masks and pronounced class imbalance hinder the development of effective segmentation models [2–5]. Metallic microstructures are imaged via optical microscopy (OM), scanning electron microscopy (SEM), electron backscatter diffraction (EBSD) and focused ion beam SEM (FIB-SEM) tomography, each producing distinct contrast and resolution that require modality-specific preprocessing and pose unique segmentation challenges [6]. Moreover, steels and alloys exhibit a wide variety of phase constituents like ferrite, pearlite, bainite, martensite, retained austenite, inclusions, etc., each occupying characteristic volume fractions that govern material properties. In the MetalDAM dataset [7], these are distilled into four classes (matrix, austenite, martensite/austenite (MA), and defect), yet abundant matrix and austenite regions far outnumber sparse MA islands and defects, exacerbating imbalance.

To address these challenges, deep learning-based segmentation methods in materials science have adopted architectures and training strategies tailored to microstructural data. U-Net variants including attention-guided U-Nets with color-space transformations [8] have been applied directly to OM and SEM images. Training tricks such as dynamic learning rates and strong augmentations [9], combined with generalist pretraining on ImageNet or domain-specific pretraining on materials datasets (e.g., MicroNet) [4], further boost accuracy. Superpixel-based unsupervised frameworks [10] and human-in-the-loop semi-supervised approaches [11] reduce annotation costs, while consistency regularization paired with contrastive learning helps exploit limited labels [12]. The MetalDAM benchmark [7] also standardizes evaluation but reveals the limits imposed by extreme class imbalance and annotation scarcity.

Generative adversarial networks (GANs) [13] have long been used to augment training sets [5,14–18] but often suffer from training instability and mode collapse. Hybrid approaches combining DCGANs with Pix2Pix produce high-resolution, realistic metal micrographs from semantic label maps [5]. More recently, denoising diffusion probabilistic models (DDPMs) [19] have demonstrated superior stability, fidelity, and robustness for scientific image synthesis, successfully generating realistic microstructural images and annotated microscopy data from rough sketches [20–23]. Two-stage diffusion approaches where image and mask synthesis occur in separate modules, have also proven effective in other fields: for example, the Segmentation-Guided Diffusion (SegGuidedDiff) model conditions each diffusion step on segmentation masks to achieve anatomically controllable medical images [24].

However, these multi-stage pipelines can introduce semantic drift and computational overhead. To streamline generation, one-stage or joint synthesis frameworks have emerged: the One-Shot Synthesis of Images and Segmentation Masks (OSMIS) framework employs a GAN-based architecture to jointly synthesize images and segmentation masks from a single annotated example, achieving high fidelity and diversity without extensive pretraining [25]; similarly, one-stage diffusion synthesis SatSynth [26] has been applied to earth observation for joint image-mask generation. Yet steel microstructures demand control over phase fractions or process rather than organ shapes, calling for a domain-specific conditioning mechanism.

In this work, we introduce a phase-fraction-controlled diffusion framework (PF-DiffSeg) that synthesizes SEM image-segmentation mask pairs in a single diffusion trajectory conditioned on a global phase-fraction vector. By explicitly specifying target phase fractions during sampling, we generate over 5,000 synthetic image-mask pairs that (i) reproduce realistic microstructural morphologies, (ii) strategically oversample rare phases to rebalance the dataset, (iii) enhance overall sample diversity, and (iv) reduce inference time. Our main contributions are:

1. A one-stage denoising diffusion model conditioned on global phase-fraction vectors for joint image and mask synthesis.

2. Scalable generation of synthetic micrograph image-mask pairs with targeted rare-phase enhancement.

3. Demonstration of improved segmentation performance, dataset diversity, and inference efficiency.

## 2. Methodology

### 2.1. Dataset

We use the publicly available MetalDAM [7] benchmark, which comprises 42 grayscale SEM micrographs of additively manufactured steel, each annotated at the pixel level into five phases: matrix, austenite, martensite/austenite (MA), precipitate and defects as shown in Fig. 1 and Table

1. For fair comparison, we reserve 5 whole micrograph images and masks as a held-out test set. To generate sufficient training samples, all micrographs and annotation masks from the train set are uniformly cropped into 400 non-overlapping tiles of size 256×256, ensuring no overlap between sets. We will also ignore precipitate class as it is out of importance in the original benchmark.

The dataset itself is heavily imbalanced with high Austenite and Matrix concentration while Martensite/Austenite and Defect present in low occurrence of below 10% of total pixels. In section 3.2. , we will discuss the sampling rare classes' phase-fraction for improving class distribution and occurrence to improve segmentation accuracy.

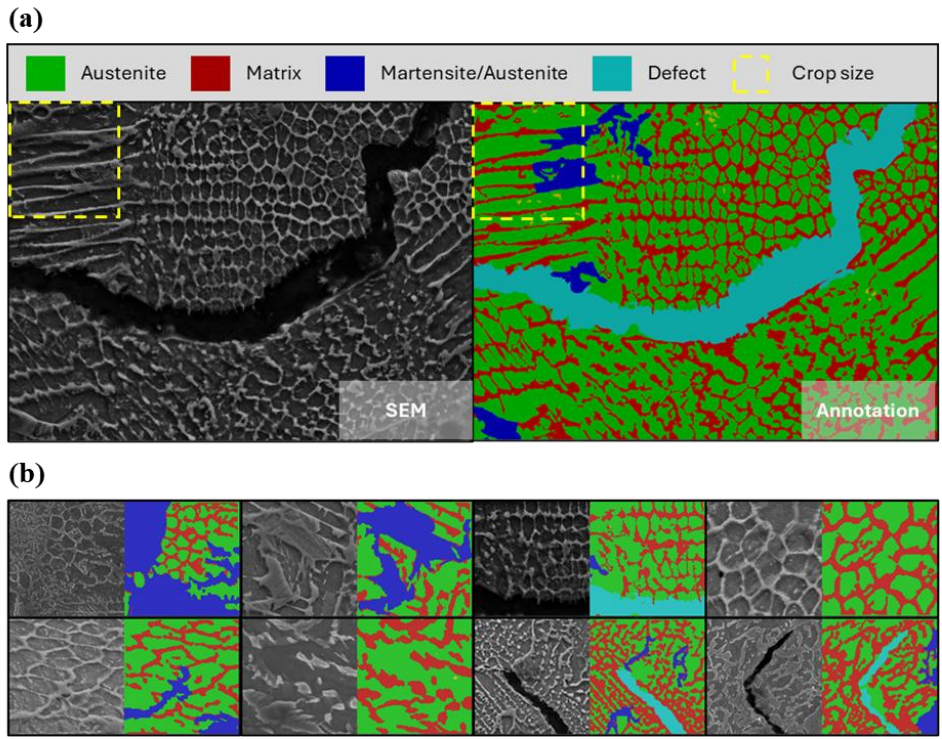

**Fig. 1.** a) Representative SEM image (left) and segmentation annotation mask (right) from MetalDAM dataset at size 1024×768, b) representative cropped image-mask pairs at size 256×256 for training.

**Table 1**

Dataset information, microconstituent classes, and ratio of MetalDAM dataset

| Index | Class  | Ratio (%) |
|-------|--------|-----------|
| 0     | Matrix | 31.86     |

| Index | Class | Ratio (%) |
|---|---|---|
| 1 | Austenite | 58.26 |
| 2 | Martensite/Austenite (MA) | 8.96 |
| 3 | Precipitate | 0.24 |
| 4 | Defect | 0.68 |

## 2.2. Phase-fraction guided image-mask pair generation

### 2.2.1. Phase-fraction guided denoising diffusion model

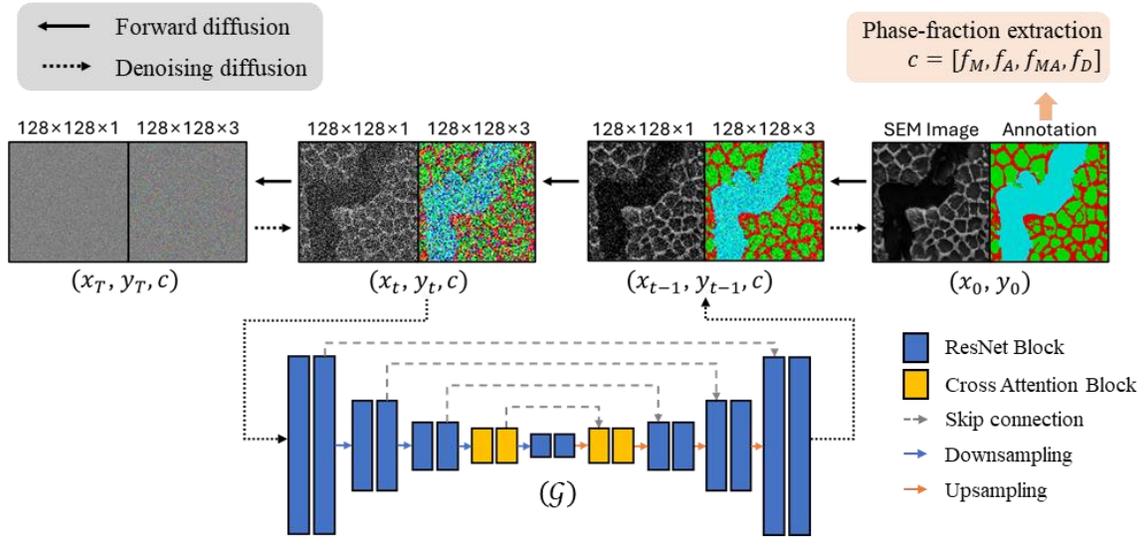

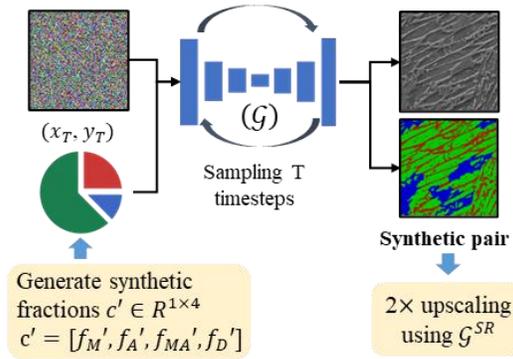

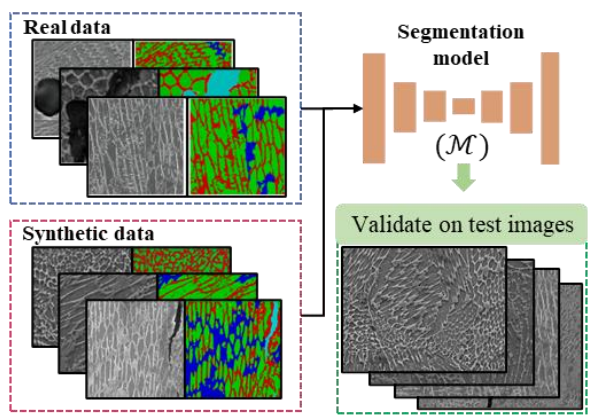

**Fig. 2.** Overview of the proposed phase-fraction controlled denoising diffusion model (PF-DiffSeg) for augmentation of microstructure image segmentation.

Our framework synthesizes paired SEM micrographs and segmentation masks in a single generative pass, leveraging a fraction-conditioned denoising diffusion model. As illustrated in Fig. 2 the method follows a simple strategy:

1. **PF-DiffSeg training:** Train a conditional denoising diffusion U-Net ($\mathcal{G}$) for image-mask pairs $(x_0, y_0)$. At each timestep $t$, $\mathcal{G}$ learns to reverse the forward noise process $(x_t, y_t)$ to $(x_{t-1}, y_{t-1})$, guided by a global phase-fraction vector $c = [f_M, f_A, f_{MA}, f_D] \in R^{1\times 4}$ with $f_i \in (0,1)$ that corresponds to the fractions of Matrix, Austenite, Martensite/Austenite and Defect within the dataset.

2. **Synthetic image-mask generation**: Sample novel conditioning vectors $c'$ from the phase-fraction space and feed pure noise $(x_T, y_T)$ plus $c'$ to $\mathcal{G}$. A single reverse-diffusion pass yields synthetic SEM images and masks $D'$ with exactly the desired phase composition.

3. **Generative data augmentation:** Merge the synthetic dataset with the real dataset to train the segmentation model $\mathcal{M}$, boosting per-class accuracy by boosting diversity and balancing rare and common phase representations.

Following denoising diffusion probabilistic models DDPM [19], we formulate the image-mask pair as a single variable $\tilde{x}_0 = [x_0, y_0] \in R^{128\times 128\times 4}$, with three channels annotation mask and one channel image are linearly normalized to the range $[-1, 1]$. The forward process applies Gaussian noise over this concatenated input:

$$q(\tilde{x}_t \mid \tilde{x}_0) = \mathcal{N}(\tilde{x}_t; \sqrt{\bar{\alpha}_t}\,\tilde{x}_0, (1-\bar{\alpha}_t)I) \qquad (1)$$

Here, $\bar{\alpha}_t = \prod_{s=1}^{t}(1-\beta_s)$ is the cumulative noise schedule up to step $t$ and $I$ is the identity matrix. The reverse process is modeled as a learnable Gaussian, where the model predicts the denoised reconstruction $\mu_\theta$:

$$p_\theta(\tilde{x}_{t-1} \mid \tilde{x}_{t-1}, c) = \mathcal{N}(\tilde{x}_{t-1}; \mu_\theta(\tilde{x}_t, t, c), \sigma_t^2 I) \qquad (2)$$

Where $\sigma_t^2$ is Gaussian noise variance, and $\mu_\theta$ is learned by denoising diffusion model $\mathcal{G}$, implemented as a six-stage U-Net with 2 ResNet blocks at each scale and multi-head cross-attention at deepest layer. The conditioning vector $c = [f_M, f_A, f_{MA}, f_D] \in R^{1\times 4}$ extracted from the annotation mask, is passed through a linear layer $R^{1\times 4} \to R^{1\times D}$, where $D$ is the time embedding $t$ dimension, and the resulting embedding is injected into each ResNet block at every scale, thereby enforcing global composition constraints throughout the denoising trajectory.

The model is trained to minimize diffusion loss:

$$\mathcal{L} = E_{\tilde{x}_0,\epsilon,t}\left[\|\epsilon - \epsilon_\theta(\tilde{x}_t, t, c)\|^2 + \lambda \|\epsilon - \epsilon_\theta(\tilde{x}_t, t, c)\|\right] \tag{3}$$

Our loss function is defined as a hybrid combination between mean squared error (MSE) in the first term and mean absolute error (L1) in the second term. $\epsilon_\theta(\tilde{x}_t, t, c)$ is the model's prediction and $\lambda$ balances the L1 term against the MSE term. Combining MSE with L1 term yields both stable, low-variance noise estimates (via squared penalty) and robustness to outliers that preserves sharp mask boundaries (via linear penalty). This hybrid loss, encouraged by [27], therefore encourages accurate overall denoising while maintaining crisp, physically meaningful phase edges.

At inference time, we draw pure Gaussian noise $\tilde{x}'_T \sim \mathcal{N}(0, I)$ and run $\mathcal{G}_\theta$ conditioned on a synthetic phase-fraction vector $c'$. The inference process follows denoising diffusion implicit model (DDIM) [28] update rule to denoise in $T = 50\ steps$.

For $t = T, T-1, \ldots, 1$, the model predicts the noise $\epsilon_\theta$, which is then used to reconstruct the clean sample $\tilde{x}'_0$:

$$\tilde{x}'_0 = \frac{1}{\sqrt{\bar{\alpha}_t}}\left(\tilde{x}'_t - \sqrt{1-\bar{\alpha}_t}\,\epsilon_\theta(\tilde{x}'_t, t, c')\right) \tag{4}$$

Then, we compute the next sample in the denoising trajectory using:

$$\tilde{x}'_{t-1} = \sqrt{\bar{\alpha}_{t-1}}\tilde{x}'_0 + \sqrt{1-\bar{\alpha}_{t-1}}\,\epsilon, \quad \epsilon \sim \mathcal{N}(0, I) \tag{5}$$

This process is repeated until $t = 1$, the recovered tensor is now $\tilde{x}'_0 = [x'_0, y'_0]$ with $x'_0 \in R^{128\times128\times1}$, $y'_0 \in R^{128\times128\times3}$ contains the new synthetic image $x'_0$ and mask $y'_0$.

Equivalently, the entire sampling process can be simplified as:

$$[x'_0, y'_0] = \mathcal{G}_\theta(z, c'), \quad z \sim \mathcal{N}(0, I) \tag{6}$$

Our one-stage formulation enables coherent synthesis of both image and mask in a fast, single 50-step DDIM [28] sampling runs, eliminating the need for separate mask prediction or multi-stage generation. This approach builds upon SegGuidedDiff [24], which applies semantic control during diffusion, and DCGAN+Pix2PixHD [5], but simplify two-stage generation to a unified paired image-mask diffusion pipeline and remove the need of a separated phase-fraction to noise mapping ML framework following Azqadan's process-aware diffusion [20] by enforcing phase fraction directly within the denoising path. The model configuration and training of PF-DiffSeg and SegGuidedDiff are shown in Table A2.

### 2.2.2. Generation of synthetic fractions for synthetic microstructure image and mask

To promote balanced phase representation, we sample synthetic phase-fraction vectors $c' = [f_M', f_A', f_{MA}', f_D']$ using a mixture-based strategy. From each real mask slice of size 256×256, we extract its phase-fraction vector $c$. We then perturb each component by adding independent Gaussian noise whose standard deviation grows with the base fraction plus a small constant:

$$c_i \leftarrow c_i + \sigma(c_i + \epsilon)\varepsilon_i \; with \; \sigma = 0.02, \epsilon = 0.01, \varepsilon_i \sim \mathcal{N}(0,1)$$

The perturbed vector is renormalized to enforce non-negativity and unit sum:

$$c_i' = \frac{c_i + \sigma(c_i + \epsilon)\varepsilon_i}{\sum_{j=1}^{K}(c_j + \sigma(c_j + \epsilon)\varepsilon_j)} \tag{7}$$

Most samples (70 %) are drawn uniformly from these perturbed real vectors; the remaining 30% are biased toward rare phases by oversampling vectors rich in martensite/austenite (20%) and defects (10%) and applying the same jitter. This yields approximately 5,000 diverse conditioning vectors $c'$, which are then used by model $\mathcal{G}$ to generate synthetic image-mask pairs.

### 2.2.3. Image-mask upscaling via super-resolution

Experiments revealed that the generative quality of our diffusion model diminishes at higher resolutions. We adopt the existing denoising U-Net model as a lightweight 2× super-resolution model $\mathcal{G}^{SR}$ inspired by [29]; this involves conditioning low-resolution 128×128 image–mask pairs to progressively refine and generate high-resolution 256×256 outputs through iterative denoising steps, enabling high-resolution training of downstream segmentation models without compromising structural fidelity.

During training, we formulate the high-resolution image-mask pair as $\tilde{x}_0 = [x_0^{HR}, y_0^{HR}] \in R^{256 \times 256 \times 4}$, and the low-resolution condition is defined as $c = [x_0^{LR}, y_0^{LR}] \in R^{128 \times 128 \times 4}$. The diffusion process is the same as in Eq. (1), (2) and (3) and the sampling inference process can be simplified as:

$$[x_0^{HR}, y_0^{HR}] = \mathcal{G}_\theta^{SR}(z, c), \quad z \sim \mathcal{N}(0, I) \tag{5}$$

This allows us to upscale synthetic images and masks generated from PF-DiffSeg model by 2 times. The training and model configuration of the super-resolution model is shown in Table **A***3*. Synthetic images and masks that will be shown in Section 3.1. are all upscaled to 256×256 and also used for training and inferring segmentation models results in Section 3.2.

### 2.3. Microstructure image segmentations

To assess the effectiveness of synthetic data generated by our fraction-conditioned diffusion model, we train multiple semantic segmentation frameworks on the MetalDAM dataset. We benchmark four established architectures U-Net [30], U-Net++ [31], LinkNet [32], and MA-Net [33], each selected to reflect diverse design strategies, including encoder-decoder symmetry, skip connections, attention mechanisms, and multi-scale feature fusion.

The segmentation models ($\mathcal{M}$) are trained under two regimes: using only real samples and using a mix of real and synthetic samples as illustrated in Fig. 2(3) with real-to-synthetic ratios of 1:1, 1:2, 1:4, 1:6, 1:8, and 1:10, which are equivalent to 5000 synthetic samples and 500 real samples (at 1:10 ratio) (only highest metrics were reported). All models share the same training pipeline, employing a ResNet-50 backbone [34], Dice loss, a batch size of 64, and 100 training epochs to ensure fair comparison. During inference, we adapt the 256×256 sliding window-inference method with 50 % overlap across each full-size SEM image, average the overlapping softmaxed outputs, and then threshold at 0.5 to obtain the final segmentation mask. We report mean intersection-over-union (MIoU), overall pixel accuracy (ACC), and per-class IoU on the held-out test set.

### 3. Results

We structure our evaluation in two stages. First, we generate synthetic image–mask pairs with PF-DiffSeg and assess their quality using standard generative metrics alongside analyses of phase-fraction distributions and embedding spaces. Second, we augment MetalDAM segmentation training with these samples and measure downstream gains in MIoU and per-class IoU. We compare PF-DiffSeg against 3 augmentation methods:

1. **Basic aug.** (Basic augmentation): Utilizing basic morphological and geometric transformations such as random brightness-contrast jittering, random rotate 90 degree, and random horizontal and vertical flips.

2. **SegGuidedDiff** [24], a two-stage diffusion framework (mask generation followed by mask-conditioned image generation), which we train on MetalDAM using the authors' hyperparameters. Then, we generate equally same 5000 synthetic samples as PF-DiffSeg and combine with real samples as augmentation.

3. **DCGAN + Pix2PixHD** [5], the two-stage GAN pipeline originally applied to MetalDAM (DCGAN for masks generation, Pix2PixHD for mask-to-image generation); we quote their reported per-class IoUs on micrographs A and B for comparison.

In what follows, Section 3.1. examines PF-DiffSeg's generative fidelity; Section 3.2. reports segmentation results when augmenting with PF-DiffSeg versus the two baselines; and Sections 3.3. analyze sample distributions, synthetic-to-real scaling effects, and 3.4. shows generation-time and performance ablations. All experiments were conducted on an Intel system with two NVIDIA RTX A100 GPUs.

## 3.1. Synthetic microstructure image and mask generation

| c = [$f_M, f_A, f_{MA}, f_D$] | [0.41, 0.45, 0.10, 0.04] | | [0.41, 0.59, 0.00, 0.00] | | [0.46, 0.36, 0.08, 0.10] | |
|---|---|---|---|---|---|---|
| | SEM | Mask | SEM | Mask | SEM | Mask |
| Synthetic | | | | | | |
| Real test | | | | | | |
| c = [$f_M, f_A, f_{MA}, f_D$] | [0.36, 0.61, 0.00, 0.00] | | [0.36, 0.56, 0.08, 0.00] | | [0.13, 0.39, 0.49, 0.00] | |
| | SEM | Mask | SEM | Mask | SEM | Mask |
| Synthetic | | | | | | |
| Real test | | | | | | |

■ Austenite   ■ Matrix   ■ Martensite/Austenite   ■ Defect

**Fig. 3.** Synthetic and real test microstructure image and mask pairs generated from PF-DiffSeg with corresponding fraction of Matrix, Martensite/Austenite, and Defect ($c = [f_M, f_A, f_{MA}, f_D]$). Each microstructure image-mask pair will be generated from a fraction condition.

We illustrate the fidelity of our conditional sampler by juxtaposing synthetic and real held-out test microstructure pairs at matching phase-fraction targets (Fig. 3). Each column lists a test phase-fraction vector $c = [f_M, f_A, f_{MA}, f_D]$ above the panels, the top row shows a PF-DiffSeg generated SEM image and its matching segmentation mask under that condition at resolution 256×256, while the bottom row presents a real test micrograph and its ground-truth mask whose measured fractions closely match $c$. Across all six settings, the generator faithfully reproduces both the global phase proportions and the detailed morphology of the corresponding real samples, demonstrating precise, semantics-aware control and high-fidelity microstructure synthesis. For example, the shape of matrix and austenite, the distribution of MA in heterogenous austenite, and the distribution of defect in matrix are replicated in the synthetic data. Although it cannot completely replicate the patterns of real data due to being conditioned on simple global phase vector, the model can generate closely matched phase composition. The visualization in Fig. 3 demonstrates that the introduction of phase-fraction control can effectively eliminate the imbalance fraction of classes. This enables the model to not only reproduce the microstructural features of real data but also provide entirely new information of microstructure with respect to phase fraction as revealed in Fig. 5.

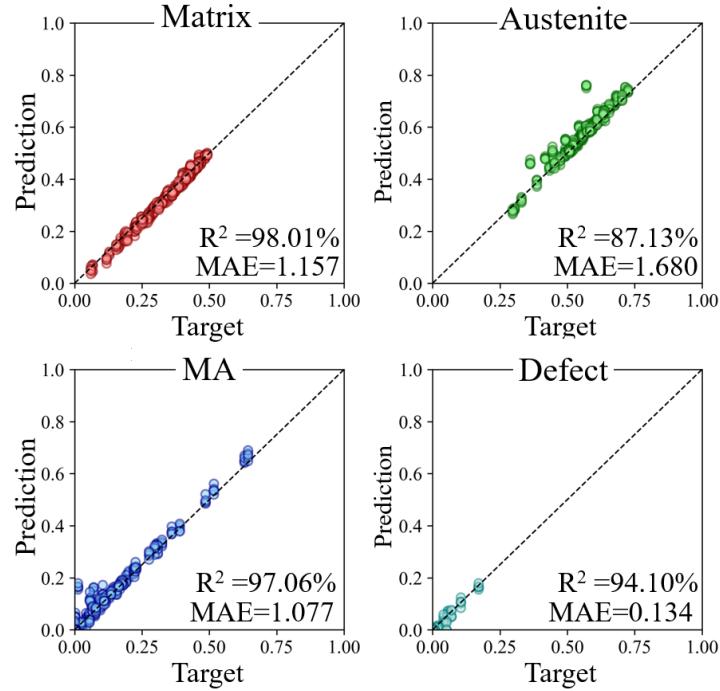

**Fig. 4.** Target fraction and predicted fraction calculated from synthetic masks of Austenite, Matrix, Martensite/Austenite, and Defect phases.

In Fig. 4, predicted fraction obtained from synthetic masks versus target phase fractions as inputted to the model are plotted for each class: Matrix ($R^2$ = 98.0%, MAE = 1.16%), Austenite ($R^2$ = 87.1%, MAE = 1.68), Martensite/Austenite ($R^2$ = 97.1%, MAE = 1.08), and Defect ($R^2$ = 94.1%, MAE = 0.13). Points tightly cluster around the diagonal in every panel, showing that the decoder recovers phase compositions with high fidelity across the full range of sampled fractions. Together, these qualitative and quantitative results confirm that our one-stage diffusion pipeline can both generate varied microstructure geometries and enforce precise phase-fraction control.

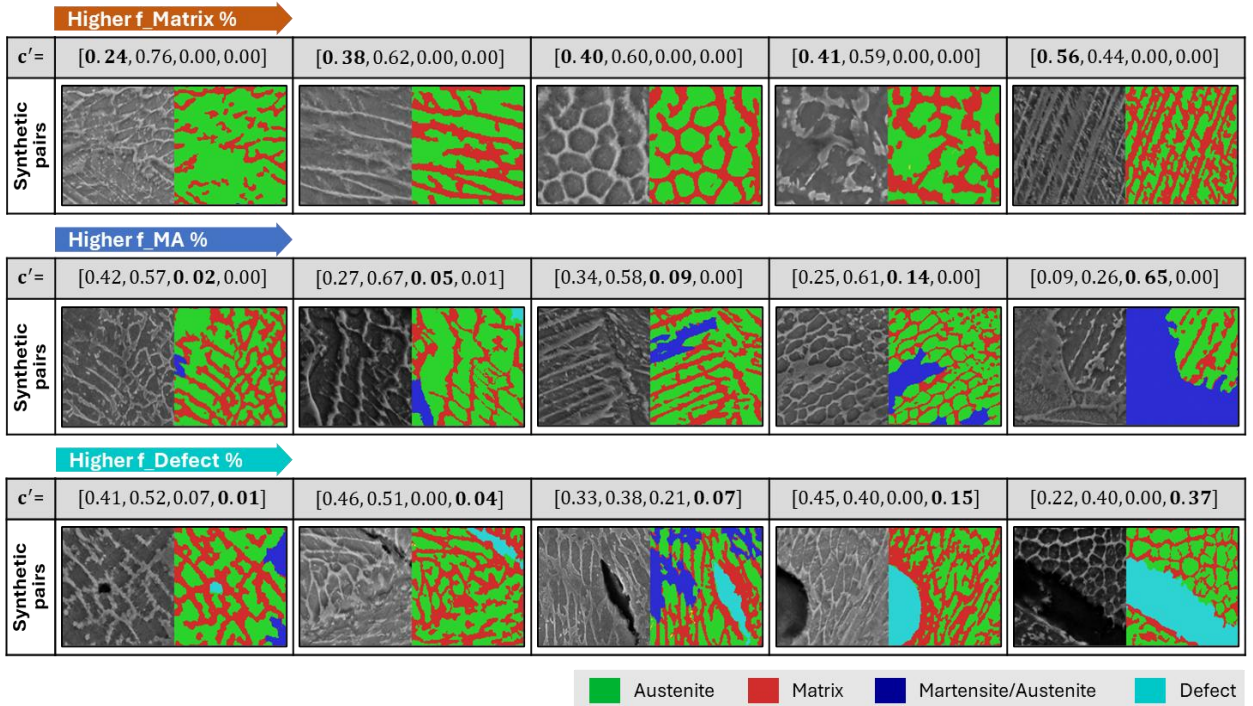

**Fig. 5.** Some samples of new synthetic microstructure image and mask pairs generated from new synthetic fractions $c'$ to combine with image and mask pairs for augmentation of segmentation models. The image-mask pairs are shown in incremental order of each class fraction to demonstrate the fraction conditioning capabilities.

**Table 2**

Evaluation metric of proposed PF-DiffSeg model for microstructure image and mask generation.

| Resolution | FID$_{image}$ (↓) | FID$_{mask}$ (↓) | IS$_{image}$ (↑) | IS$_{mask}$ (↑) | Pre (↑) | Rec (↑) |
|---|---|---|---|---|---|---|
| 128x128 | 61.61 | 68.85 | 3.686 | 2.087 | 0.593 | 0.778 |
| 256x256 | 74.52 | 34.77 | 3.695 | 2.362 | 0.371 | 0.706 |

To assess the fidelity and diversity of synthetic microstructure image-mask pairs, we employ Fréchet Inception Distance (FID) and Inception Score (IS) for both image and mask, calculating each metric over 5,000 synthetic samples against the 500 real samples. Additionally, we report precision (Pre) and recall (Rec) of images to characterize the coverage of the learned distribution. Results for two resolutions 128×128 and 256×256 (upscaled via super-resolution) are summarized in Table 2. At 128×128 resolution, evaluation metrics FID$_{image}$, FID$_{mask}$, Pre, and Rec attest to the baseline realism and distribution coverage of our synthetic pairs based on analysis of generated synthetic samples. Upscaling to 256×256 via super-resolution further reduces $\text{FID}_{\text{mask}}$ and raises mask $\text{IS}_{\text{image}}$, with only marginal shifts in image metrics, showing that the super-resolution step preserves sample fidelity and diversity. A small ablation on how loss selection affect generation fidelity is shown in Table A1.

## 3.2. Microstructure segmentation performances

To evaluate the utility of synthetic data, we train several canonical segmentation architectures UNet [30], UNet++ [31], LinkNet [32], and MANet [33], each representing different design philosophies, such as encoder-decoder symmetry, attention mechanisms, and multiscale feature integration. This architectural diversity ensures that observed performance trends are robust and not confined to a particular model type. Our evaluation metrics are intersection over union (IoU), mean intersection over union (MIoU) and pixel accuracy (ACC).

**Table 3**

Segmentation metrics (MIoU%, ACC%) on MetalDAM among various segmentation frameworks with different augmentation methods.

| | | U-Net | | U-Net++ | | LinkNet | | MANet | |
|---|---|---|---|---|---|---|---|---|---|
| Resolution | Method | MIoU (↑) | ACC (↑) | MIoU (↑) | ACC (↑) | MIoU (↑) | ACC (↑) | MIoU (↑) | ACC (↑) |
| 128×128 | Basic aug. | 63.35 | 82.44 | 62.70 | 82.34 | 61.32 | 81.34 | 63.34 | 83.27 |
| | SegGuidedDiff | 62.32 | 82.36 | 63.04 | 82.65 | 62.52 | 82.87 | 62.86 | 82.17 |
| | **PF-DiffSeg** | **66.28** | **84.11** | **66.11** | **84.30** | **66.62** | **84.61** | **65.04** | **83.75** |
| 256×256 | Basic aug. | 66.18 | 85.01 | 66.77 | 84.79 | 65.28 | 84.93 | 66.79 | 84.20 |
| | SegGuidedDiff | 65.53 | 76.97 | 68.58 | 85.23 | 67.12 | 84.89 | 67.08 | 85.01 |
| | **PF-DiffSeg** | **69.17** | **86.13** | **70.13** | **86.38** | **70.47** | **86.39** | **69.49** | **86.20** |

s

As shown in Table 3, we benchmark three training protocols: Basic augmentation, SegGuidedDiff, and our single-stage PF-DiffSeg (boosted MA and defect) across four segmentation backbones at both 128×128 and 256×256 upscaled resolutions. PF-DiffSeg consistently yields the highest MIoU and ACC, boosting average MIoU by about 3-5 % over Basic aug. and 1.5-3.5 % over SegGuidedDiff. Fig. 6 further breaks these gains down on a per-class basis (Matrix, Austenite, MA,

Defect), comparing PF-DiffSeg (orange) against Basic aug. (blue), SegGuidedDiff (purple), and PF-DiffSeg with targeted MA/Defect oversampling (red). All experiments are conducted 5 times with different random seeds. While PF-DiffSeg alone raises MA IoU from ∼41 % to ∼55 % and Defect IoU from ∼50 % to ∼70 %, the boosted-phase variant pushes MA above 60 % and Defect beyond 80 %, without sacrificing performance on the abundant Matrix and Austenite classes. This demonstrates that our phase-fraction conditioning, especially when combined with minority-phase oversampling effectively balances class representation and sharpens segmentation of rare microstructural phases.

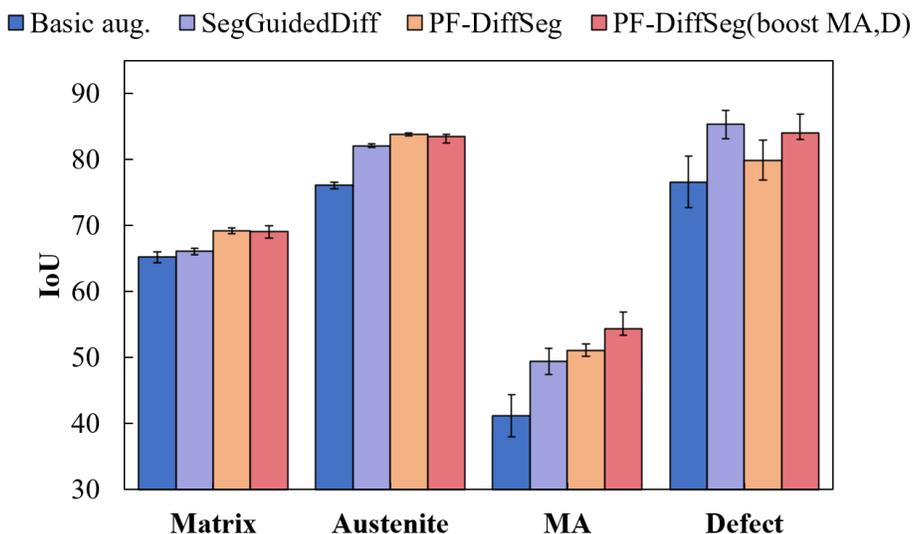

**Fig. 6.** Per-class IOU% between Basic augmentation, synthetic methods SegGuidedDiff, PF-DiffSeg and also PF-DiffSeg with boosted minority classes (Martensite/Austenite) MA and Defect.

Fig. 7 contrasts model outputs on 3 test micrographs. As indicated by white dash box in Fig. 7, the model trained with basic augmentation misses several slender MA islands (blue) falsely identified them as Matrix or vice versa, whereas model augmented with PF-DiffSeg delineates these regions accurately, mirroring the ground truth. In Fig. 7(a), a predominantly martensitic matrix with fine austenite laths: Basic aug. under-segments slender austenite regions (green) and over-predicts matrix (red), whereas PF-DiffSeg recovers these narrow austenite bands. In Fig. 7(b), high MA-fraction micrograph: Basic aug. fragments continuous MA islands and mislabels small cavities as matrix, while PF-DiffSeg produces cohesive MA regions and correct cavity delineation. In Fig. 7(c), dominant austenite network with sparse MA: Basic aug. misses fine MA islands and sometimes splits austenite bands; in contrast, PF-DiffSeg better captures slender MA structures and preserves continuous austenite regions. The results demonstrate that the proposed approach with PF-DiffSeg shows superior performance compared to the conventional method with basic augmentation.

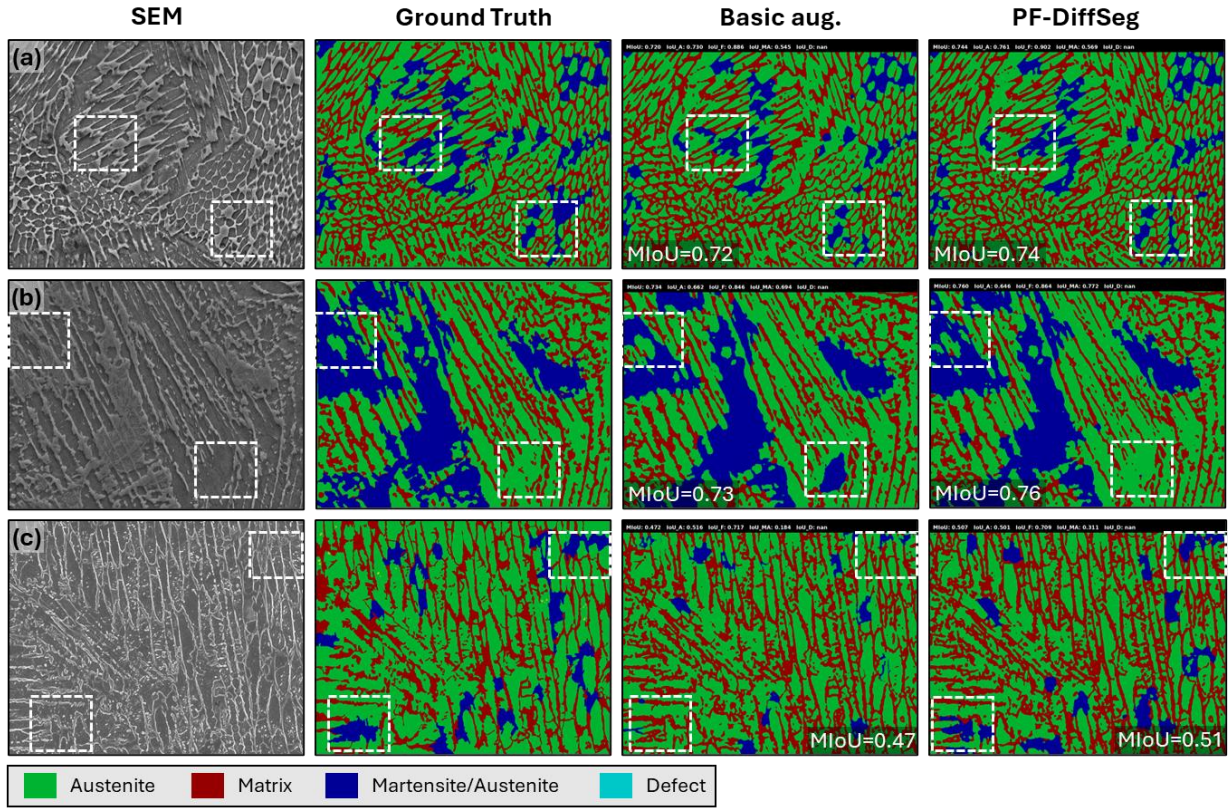

**Fig. 7.** Qualitative comparison of segmentation predictions for 3 representative test micrographs among Ground Truth (Annotation mask) of Basic aug. and PF-DiffSeg.

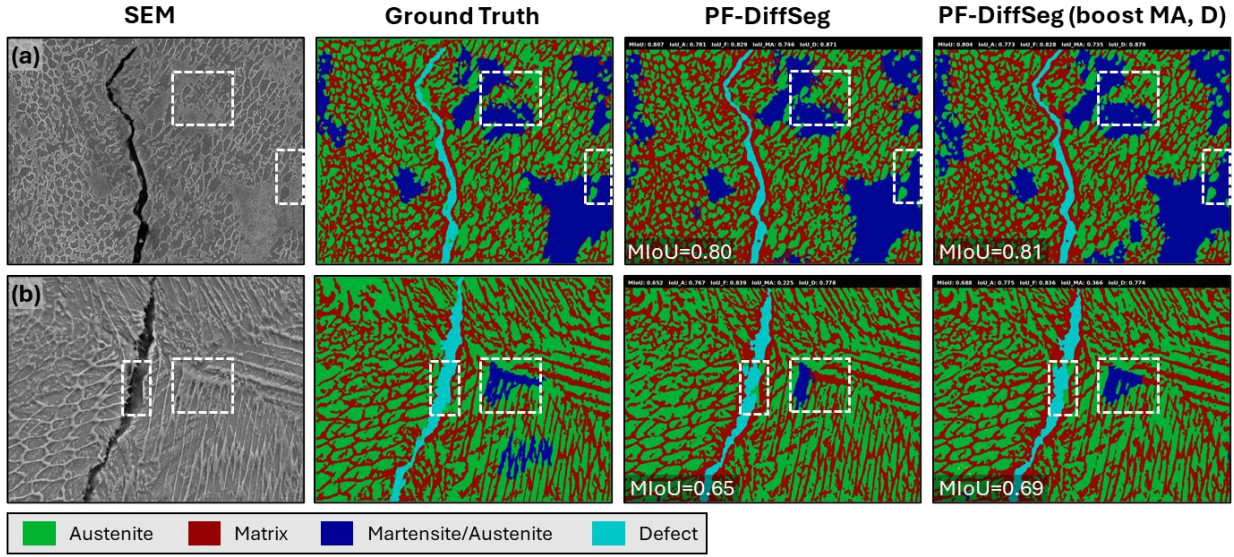

**Fig. 8.** Qualitative comparison of segmentation predictions of 2 test micrographs among Ground Truth, PF-DiffSeg augmented model and PF-DiffSeg with boosted rare MA and defect phase.

We show PF-DiffSeg's outputs with and without MA and defect phase boosting on two remaining test micrographs in Fig. 8. In Fig. 8(a), the unboosted model already segments most MA (blue) and defects (cyan) well but misses fine MA near the crack to the right; boosting recovers those islands but also misidentified the matrix patch as MA making it slightly worse. In Fig. 8(b), vanilla PF-DiffSeg not only fragments defect streaks and thin MA veins but even inserts a spurious austenite spike (green) inside a defect pocket (cyan), boosting both restores coherent MA/defect regions and removes that artefact. Rare-phase boosting thus sharpens PF-DiffSeg's delineation of slender MA filaments and small cavities.

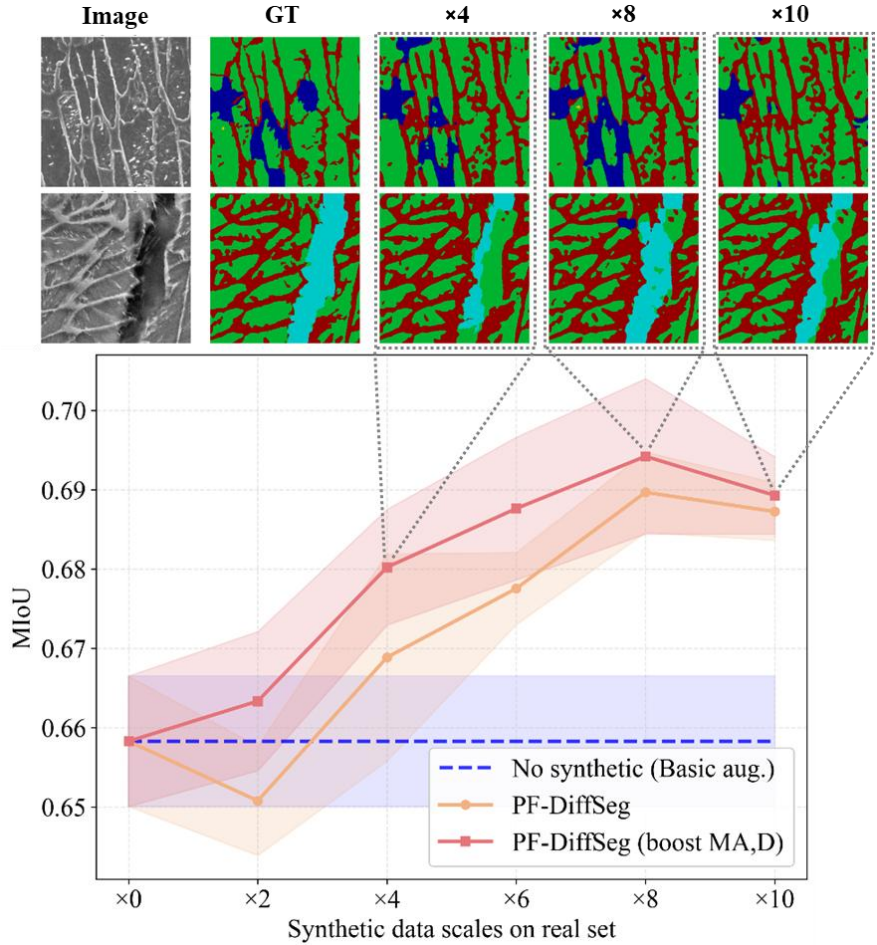

**Fig. 9.** Synthetic data scaling behavior: scaling synthetic data volume w.r.t real data affects segmentation performance (MIoU).

We varied the synthetic data volume from ×0 (no synthetic) up to ×10 relative to the real set and measured segmentation performance (MIoU). In Fig. 9, the dashed blue line marks the baseline using only basic augmentations; the orange curve represents PF-DiffSeg without rare-phase boosting, and the red curve includes 20% MA and 10% defect boosting. Both PF-DiffSeg variants begin showing substantial gains at ≥ ×4 synthetic data, with performance peaking near ×8 before plateauing due to domain overfitting. Notably, the boosted model consistently outperforms the non-boosted variant across all scales, especially in the ×2 to ×6 range, confirming that oversampling rare phases yields greater and more reliable benefits than simply increasing volume. Prediction visualizations from each model (×4, ×8 and ×10) are shown above, illustrating progressive improvements and degradation in segmenting rare-phase regions (e.g., MA islands and defect cavities) as synthetic volume increases. Shaded bands denote standard deviation over five runs, and bold lines indicate mean MIoU.

**Table 4**

Comparison with other generative method DCGAN+Pix2PixHD [5] on MetalDAM dataset

| Val. data | Method | MIoU | IoU_A | IoU_M | IoU_MA |
|---|---|---|---|---|---|
| A | DCGAN+Pix2PixHD [5] | 0.638 | 0.739 | 0.482 | 0.672 |
| | **Ours** | **0.760** | **0.864** | **0.646** | **0.772** |
| B | DCGAN+Pix2PixHD [5] | 0.541 | 0.692 | 0.507 | 0.426 |
| | **Ours** | **0.744** | **0.902** | **0.761** | **0.569** |

As part of the comparisons, we also benchmarked our approach against a generative adversarial augmentation counterpart DCGAN + Pix2PixHD [5] on 2 validation micrographs A and B with IoU scaled to 0 and 1. As shown in Table 4, our method exhibits clear superiority in MIoU and per-class IoU, especially for challenging phases. These findings suggest that diffusion models, when conditioned appropriately, are better suited for microstructural diversity modeling than adversarial counterparts. We found that our gains likely come from the properly conditioned denoising diffusion model, which has been proved to be more effective against GAN counterparts [19] and the effective test-time inference strategy discussed in section 2.3. Our segmentation results of val. data A and B are shown in Fig. 7(a and b).

### 3.3. Detailed analysis of synthetic generations

Fig. 10 compares, for the three key phases: Matrix, MA, and Defect, the normalized phase-fraction histograms produced by unconditional SegGuidedDiff versus our phase-fraction–conditioned PF-DiffSeg with targeted MA/Defect boosting. Under SegGuidedDiff (Fig. 10(b)), the synthetic distributions (red) cluster around moderate fractions and severely under-represent both the high-fraction tails and the lowest extremes seen in the real data (blue), particularly for MA and Defect. In contrast, PF-DiffSeg (Fig. 10(a)) not only aligns its red bars with the real histograms across the main mode of Matrix but thanks to our phase-fraction conditioning plus minority-phase oversampling, successfully populates the dashed-box regions at the high ends of the MA and Defect curves. This demonstrates that PF-DiffSeg can reproduce real data composition statistics while maintaining sufficient coverage of rare, high-fraction microstructural phases.

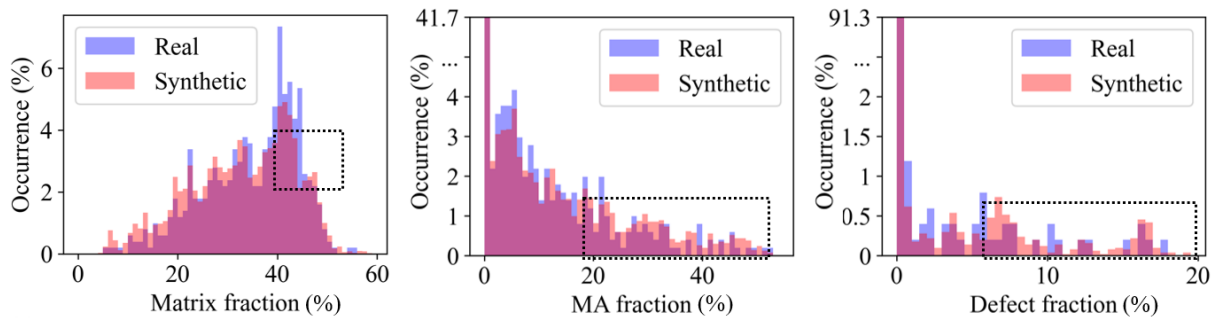

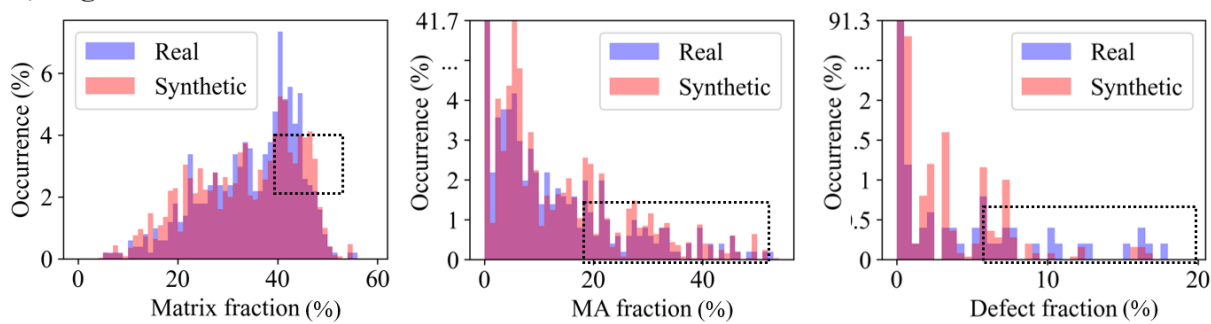

**Fig. 10.** Phase-fraction comparison for Matrix, Austenite, MA, and Defect between (a) unconditional generation SegGuidedDiff and (b) phase-fraction conditional PF-DiffSeg (synthetic data volume is ×10 larger than real data but normalized to percentages for comparability).

Qualitative and quantitative analysis of PF-DiffSeg's generated images and masks are shown in Fig. 11(a and b) show T-SNE projections of the learned latent embeddings for synthetic microstructure images and masks, respectively: synthetic samples (red) closely overlap with real training (blue) and test (green) embeddings, indicating that the model faithfully captures the intrinsic feature distribution. However, majority of synthetic data lies beyond real data distribution in the latent space, this explains the plateauing effect observed in Fig. 9. Analyzing the class ratio of synthetic dataset shows that by boosting MA and Defect fractions, the ratio MA and Defect doubled from the real dataset as seen in Fig. 12(a). This balancing strategy helps mitigate the underrepresentation of minority phases with MA and defect fractions percentage doubled while matrix and austenite fraction slightly reduce to compensate. Additionally, the 4-D synthetic fraction vectors $c'$ and real vectors $c$ are reduced to 2-D using Principal Component Analysis (PCA) and visualized using T-SNE in Fig. 12(b).

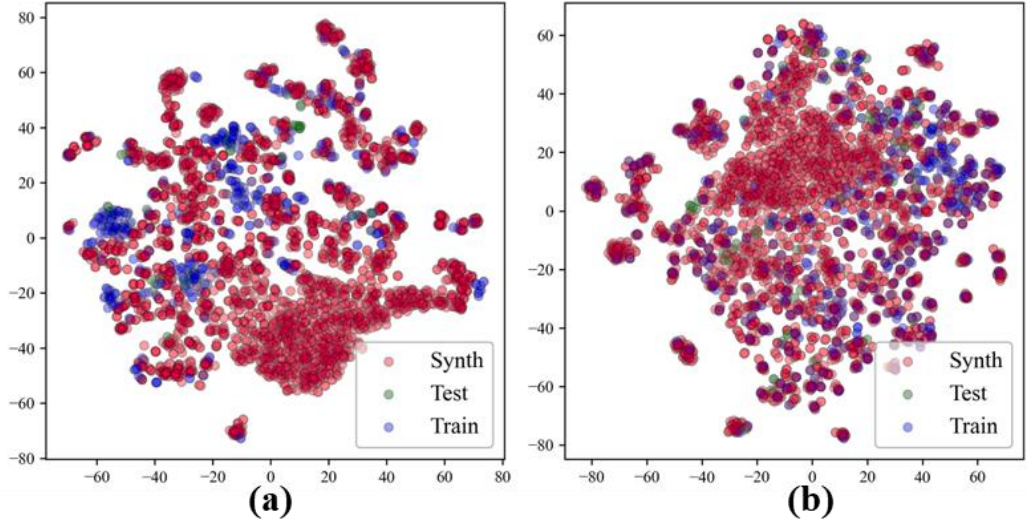

**Fig. 11.** T-SNE visualization of synthetic samples (a) synthetic image and (b) mask embeddings on 2D plane (synthetic data volume is ×10 larger than real data but normalized to percentages for comparability).

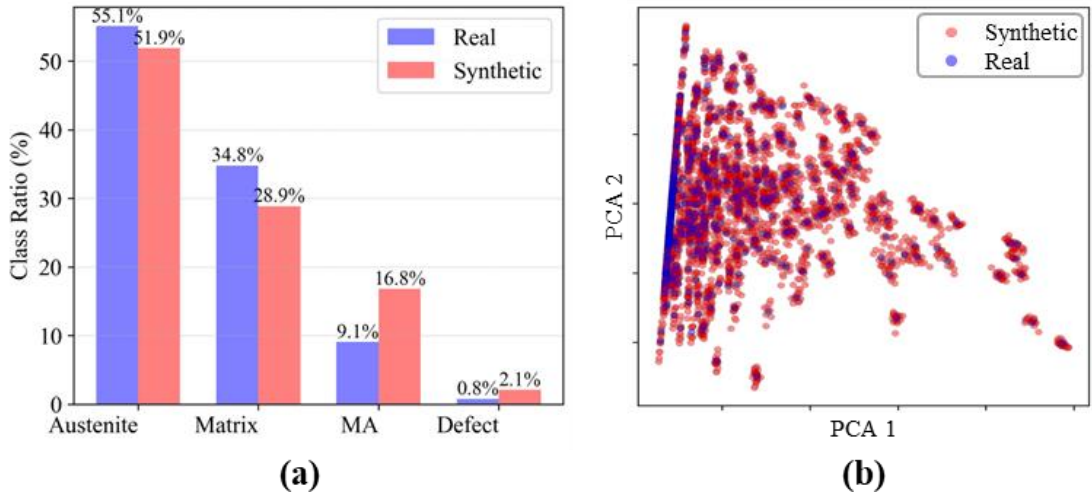

**Fig. 12.** (a) Class ratio analysis between real and synthetic dataset, (b) Visualization of synthetic and real phase fraction on 2D latent space.

We demonstrated 2× upscaling results using super-resolution model $\mathcal{G}^{SR}$ in Fig. 13. The upscaled images reveal clearer structures with enhanced sharpness and phase edges on various synthetic samples. This shows that DDPM models, besides being conditioned with phase-fraction information, can be also conditioned with high-resolution images to use as an image upscaler. This is essential for metallograph image synthesis tasks because generative models are sometimes too demanding to work with, and generating smaller resolution samples then upscale them to higher resolution could be a more reliable approach.

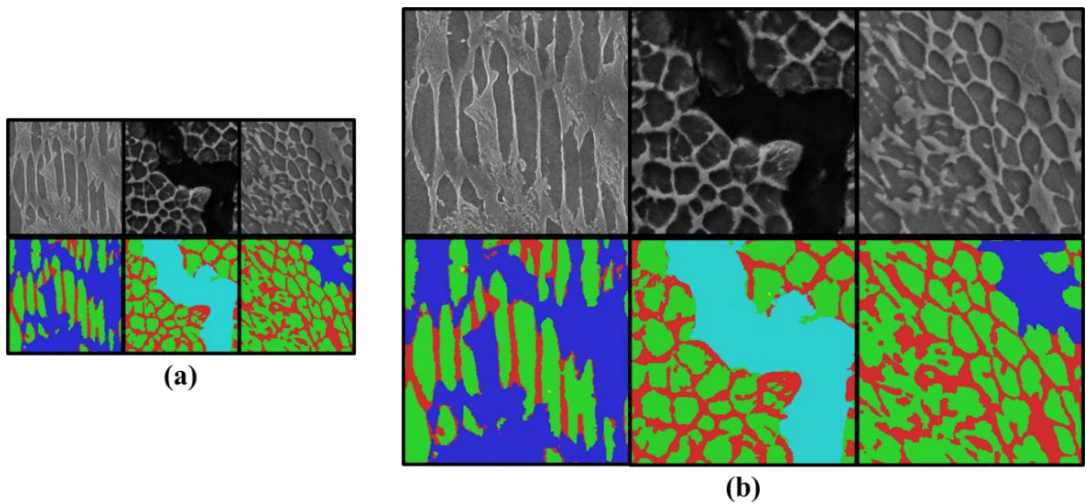

**Fig. 13.** (a) Synthetic SEM image (top) and mask (bottom) at resolution 128×128, (b) 256×256 upscaled image and mask using super-resolution model $\mathcal{G}^{SR}$.

## 3.4. Comparison of generation strategies

**Table 5**

Ablated comparison on Single Image Generation Time (SIGT) and MIoU segmentation metric.

| Method | Generation strategy | SIGT (sec) | MIoU | MIoU (val. A) | MIoU (val. B) |
|---|---|---|---|---|---|
| SegGuidedDiff | Mask | 1.20 | 68.58 | - | - |
|  | Image | 1.21 |  |  |  |
| DCGAN+Pix2PixHD [5] | Mask | - | - | 63.80 | 54.10 |
|  | Image | - |  |  |  |
| **PF-DiffSeg** | Image + Mask | **2.01** | 70.42 | 77.20 | 71.20 |

A more detailed analysis of Table 5 reveals three key insights:

- **Two-stage overhead:** SegGuidedDiff and DCGAN+Pix2PixHD [5] must be applied twice. For SegGuidedDiff, generating the mask (1.20s) and translating to image (1.21s) for a total of 2.41s per complete image-mask pair not to mention the training time of two diffusion models with DDIM sampling. By contrast, our PF-DiffSeg produces both outputs in a single 2.01s pass, saving ~17% of the generation time and only needs to be trained once.
- **Alignment gains:** Our synthetic data augmentation achieves higher segmentation performance than SegGuidedDiff, with an MIoU of 70.42% vs. 68.58%, and also outperforms prior GAN-based [5]. These gains stem not only from the joint generation of image and mask, which helps preserve spatial coherence, but also from the model's effective direct phase fraction conditioning strategy, enabling better alignment and consistency in the synthesized pairs.
- **Practical trade-off:** For systems with limited computing resources or latency constraints, the slightly simpler two-stage denoising diffusion (mask generation and mask-to-image translation) or even generative adversarial training (DCGAN+Pix2pixHD [5]) may be preferable. However, on sufficiently provisioned hardware, our one-stage approach offers both higher accuracy and lower end-to-end latency for full image-mask generation, making it attractive for high-throughput metallographic pipelines.

## 4. Discussion

The empirical study confirms three key advantages of fraction-controlled, one-stage diffusion for steel micrograph data augmentation for segmentation (PF-DiffSeg).

1. The comparison in Table 5 shows that decoupling mask and image generation not only doubles inference time but also erodes MIoU by almost two percentage points. This gap stems from spatial drift introduced when the mask is generated first and subsequently translated into a photorealistic SEM texture, small contour perturbations propagate into sizeable pixel-level errors after style transfer. One-stage denoising resolves this issue by treating the image-mask pair as a single high-dimensional sample, thereby enforcing alignment at every timestep. We also provided evolution of test micrograph image-mask pair during training of the diffusion model in Fig. A1.
2. Our conditioning vector is a coarse yet physically interpretable global descriptor. During training, the denoiser learns to associate specific fraction patterns with characteristic morphologies (e.g., sparse MA islands embedded in ferrite). At sampling time, the model can be steered towards otherwise under-represented compositions, effectively re-weighting the data distribution without handcrafted oversampling. The strong per-class gains for MA and defects (Fig. 6) corroborates this interpretation. Our methods seem to be effective when the synthetic presents in large volume by around 6-10 times the volume of real set as discussed in scaling behavior in Fig. 8. Our scaling behavior is in line with [26].

3. While the MetalDAM benchmark covers a realistic range of AM steel microstructures, its limited amount of data may not capture the full variability found in industrial alloys. The framework is also supposed to work with various metal alloys with complex or imbalance phase distribution and different imaging techniques like OM, SEM, TEM or EBSD. Extending the framework to large image size microscopy images or EBSD, and 3D tomography volumes will require memory-efficient transformers or Latent Diffusion models. Moreover, fraction vectors encode only global composition; incorporating shape statistics (e.g., particle elongation, grain size) or spatial priors could yield finer control. Finally, domain adaptation and self-distillation techniques could further bridge the gap between synthetic and real domains. Finally, integrating domain-adaptation or self-distillation techniques could further close the gap between synthetic and real-world data, and applying our method to automate not only segmentation but also downstream materials discovery and quality-control testing pipelines presents a promising avenue.

## 5. Conclusion

We have presented a fraction-conditioned, one-stage denoising diffusion framework that jointly generates steel microstructure images and segmentation masks in a single inference pass (PF-DiffSeg) that can boost performance of steel microstructure segmentation task. Compared to standard augmentation and two-stage baselines, our method boosts MIoU by up to 4% and per-class IoU for rare phases by over 10%, while reducing end-to-end generation time by 17%. By embedding domain-specific composition constraints directly into the generative path, we deliver high-fidelity, physically consistent synthetic samples that strengthen segmentation performance, particularly for under-represented microconstituents, without additional manual tuning. This scalable, efficient augmentation strategy can partly reduce manual annotation for analysis systems and can be readily integrated into automated materials testing and discovery systems, accelerating high-throughput metallographic analysis. Future extensions will target high-resolution 3D imaging modalities, Latent Diffusion style [35] image encoding for more lightweight modeling and adaptive conditioning mechanisms to further enhance applicability across diverse materials characterization tasks.

**Code Availability**

Source code, training scripts are available at https://github.com/namKolorfuL/PF-DiffSeg.

**Data Availability**

This paper uses MetalDAM dataset, which is publicly available through its author's website.

**Competing Interests**

The authors declare no competing interests.

**Acknowledgments**

This research was supported by Nano and Materials R&D program (RS-2024-00451579) through the Korea Science and Engineering Foundation and Basic Research Program of Korea Institute of Materials Science (PNKA640) funded by the Ministry of Education, Science and Technology.

**Appendix**

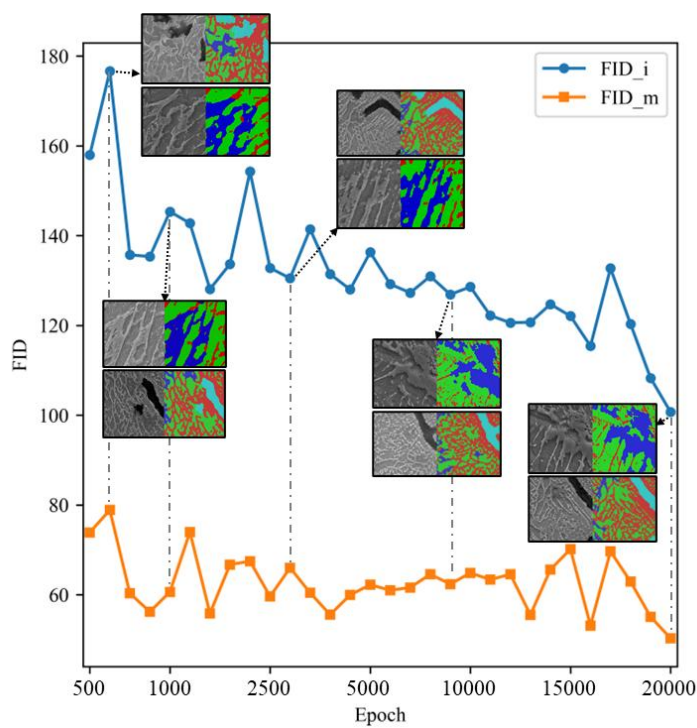

**Fig. A1.** FID convergence curves of fraction guided image-mask generation model over 20000 epoch training. Visualization images and masks are sampled from single phase-fraction vector from test set.

Fig. A1 traces the Fréchet Inception Distance (FID) for images (FID_i, blue) and masks (FID_m, orange) over 20000 training epochs of our phase-fraction–guided diffusion model. Both curves drop sharply in the first 1000 epochs, reflecting rapid learning of coarse structure, then enter a series of damped oscillations as the network fine-tunes textures and boundary details. The consistently lower FID_m values indicate that mask synthesis converges faster and is less sensitive to high-frequency noise than image synthesis. Dashed vertical lines mark epochs at which example image-mask pairs are shown early (≈1000 epochs) snapshots exhibit coarse phase delineation; mid-training (≈2500–5000 epochs) captures finer austenite and MA morphologies; later (≈15000 epochs) reveals clearer defect boundaries; and by 20000 epochs the model produces realistic microstructure patterns with accurate phase masks. After ≈10000 epochs, both FID curves plateau, confirming stable long-run convergence.

**Table A1.** Ablation study on how loss functions affect generative quality at resolution 128×128

| Training objective | $FID_{image}$ (↓) | $FID_{mask}$ (↓) | $IS_{image}$ (↑) | $IS_{mask}$ (↑) |
|---|---|---|---|---|
| L1 only | 78.23 | 50.48 | 2.730 | 1.765 |
| MSE only (λ=0) | 109.0 | 70.37 | 2.748 | 1.605 |
| MSE + L1 (λ=0.5) | 92.89 | 64.21 | 2.864 | 1.760 |
| **MSE + L1 (λ=1)** | **61.61** | **68.85** | **3.686** | **2.087** |

**Table A2.** PF-DiffSeg and SegGuidedDiff model and training configuration

| | **PF-DiffSeg** | **SegGuidedDiff (Mask/Image)** |
|---|---|---|
| **Model configuration** | | |
| Layer | ResNet + Self/Cross-attention | ResNet + Self-attention |
| Number of layers | 2 | 2 / 2 |
| Layer depths | 128, 128, 256, 256, 512, 512 | 128, 128, 256, 256, 512, 512 |
| Condition size | 1×4 (phase-fraction) | None / 128×128×3 (mask) |
| Input/Output channel | 128×128×4 (image + mask) | 128×128×3/128×128×1 |
| **Hyperparameters** | | |
| Loss function | MSE+L1 | MSE |
| Learning rate | 0.0001 | 0.0001 |
| Scheduler | Cosine | Cosine |
| Optimizer | AdamW | AdamW |
| Epoch | 20000 | 10000 / 10000 |

**Table A3.** Image super-resolution model training and configuration

| **Model configuration** | **Parameters** |
|---|---|
| Layer | ResNet + Self-attention |
| Number of layers | 2 |
| Layer depths | 128, 128, 256, 256, 512, 512 |
| Input/Output channel | 128×128×8/256×256×4 |

| Hyperparameters | |
|---|---|
| Loss function | MSE |
| Learning rate | 0.0001 |
| Scheduler | Cosine |
| Optimizer | AdamW |
| Epoch | 5000 |